\documentclass[conference]{IEEEtran}
\IEEEoverridecommandlockouts

\usepackage{cite}
\usepackage{amsmath,amssymb,amsfonts}
\usepackage{algorithmic}
\usepackage{graphicx}
\usepackage{textcomp}
\usepackage{xcolor}
\usepackage{booktabs}
\usepackage{multirow}
\usepackage[table]{xcolor}
\def\BibTeX{{\rm B\kern-.05em{\sc i\kern-.025em b}\kern-.08em
    T\kern-.1667em\lower.7ex\hbox{E}\kern-.125emX}}
\begin{document}

\title{Towards Faithful Sentimental Image Captioning via Evidence-Aware Multi-Agent Reasoning
}

\author{\IEEEauthorblockN{1\textsuperscript{st} Tiecheng Cai}
\IEEEauthorblockA{\textit{College of Computer and Data Science,} \\
\textit{Fuzhou University}\\
Fuzhou, China \\
210310002@fzu.edu.cn}
\and
\IEEEauthorblockN{2\textsuperscript{nd} Zexian Yang*}
\IEEEauthorblockA{\textit{College of Computer and Data Science,} \\
\textit{Fuzhou University}\\
Fuzhou, China \\
yangzexian@fzu.edu.cn}
\and
\IEEEauthorblockN{3\textsuperscript{rd} Chao Chen}
\IEEEauthorblockA{\textit{School of Computer Science and } \\
\textit{Technology, Harbin Institute of Technology} \\
Shenzhen, China \\
cha01nbox@gmail.com}
\and
\IEEEauthorblockN{4\textsuperscript{th} Shanshan Lin}
\IEEEauthorblockA{\textit{College of Computer and Data Science,} \\
\textit{Fuzhou University}\\
Fuzhou, China \\
231016006@fzu.edu.cn}
\and
\IEEEauthorblockN{5\textsuperscript{th} Xiangwen Liao}
\IEEEauthorblockA{\textit{College of Computer and Data Science,} \\
\textit{Fuzhou University}\\
Fuzhou, China \\
liaoxw@fzu.edu.cn}
}
\maketitle

\begin{abstract}
Sentimental Image Captioning (SIC) requires balancing emotional expression with visual fidelity. 
Existing methods often struggle with this trade-off, leading to hallucinations due to insufficient local grounding and the lack of sentimental verification mechanisms. 
To address these limitations, we propose SEA-Cap, a Sentiment-Evidence-Aware Multi-Agent System for faithful and evidence-grounded sentimental image captioning. SEA-Cap incorporates a Sentiment Evidence Miner that extracts structured, local affective cues to shift sentiment control from global attributes to verifiable object-level evidence. Leveraging this evidence, our framework orchestrates a collaborative workflow where a Generator, Hallucination Checker, and Arbitrator iteratively refine captions via a shared blackboard. By explicitly auditing generated content against mined visual evidence, SEA-Cap ensures both sentiment accuracy and factual consistency. Extensive experiments on two benchmark datasets demonstrate that SEA-Cap effectively mitigates hallucinations and achieves state-of-the-art performance.
\end{abstract}


\section{Introduction}
\label{sec:intro}

Image captioning (IC) aims to produce a factual, objective description of visual content \cite{IC}. In contrast, sentimental image captioning (SIC) requires a model to understand an image from different sentimental perspectives and to generate a caption consisting of the target sentiment \cite{Senticap}. Recent researchers have endeavored to develop a model that can harmonize sentiment precision with visual fidelity \cite{Senticap,Stylenet,Semstyle}. Nevertheless, these sentiment-oriented methods are designed to operate under a single constraint. Practical requirements, such as a strict word budget, frequently cause these models to suffer from sentiment drift or a decline in descriptive fidelity.

In order to address these challenges, recent work has focused on managing the trade-off between caption length and expressive granularity. This line of work recognizes shorter captions tend to emphasize high-level visual cues, whereas longer ones facilitate richer affective details but increase the risk of introducing ambiguous or mixed sentiments \cite{sweller1988cognitive}. Instead of relying on standalone Large Language Models (LLMs) or Multimodal LLMs (MLLMs) where prompt-based steering often yields inconsistent sentiment control \cite{Concap, Conzic}, the Multi-Agent System (MAS) framework has been explored for image captioning to enhance diversity, factual consistency, and contextual alignment \cite{bai2025power, lee2025toward}. Along this line, the training-free method IE-MAS \cite{cai2025ie} integrates internal representation-level steering with external multi-agent collaboration to improve controllability and coordination under multiple objectives. However, achieving seamless coordination remains a significant hurdle due to inherent conflicts among heterogeneous objectives, which may further induce hallucinations \cite{yang2019image,yang2024pedestrian}.

\begin{figure}[t]
\centerline{\includegraphics[width=0.5\textwidth]{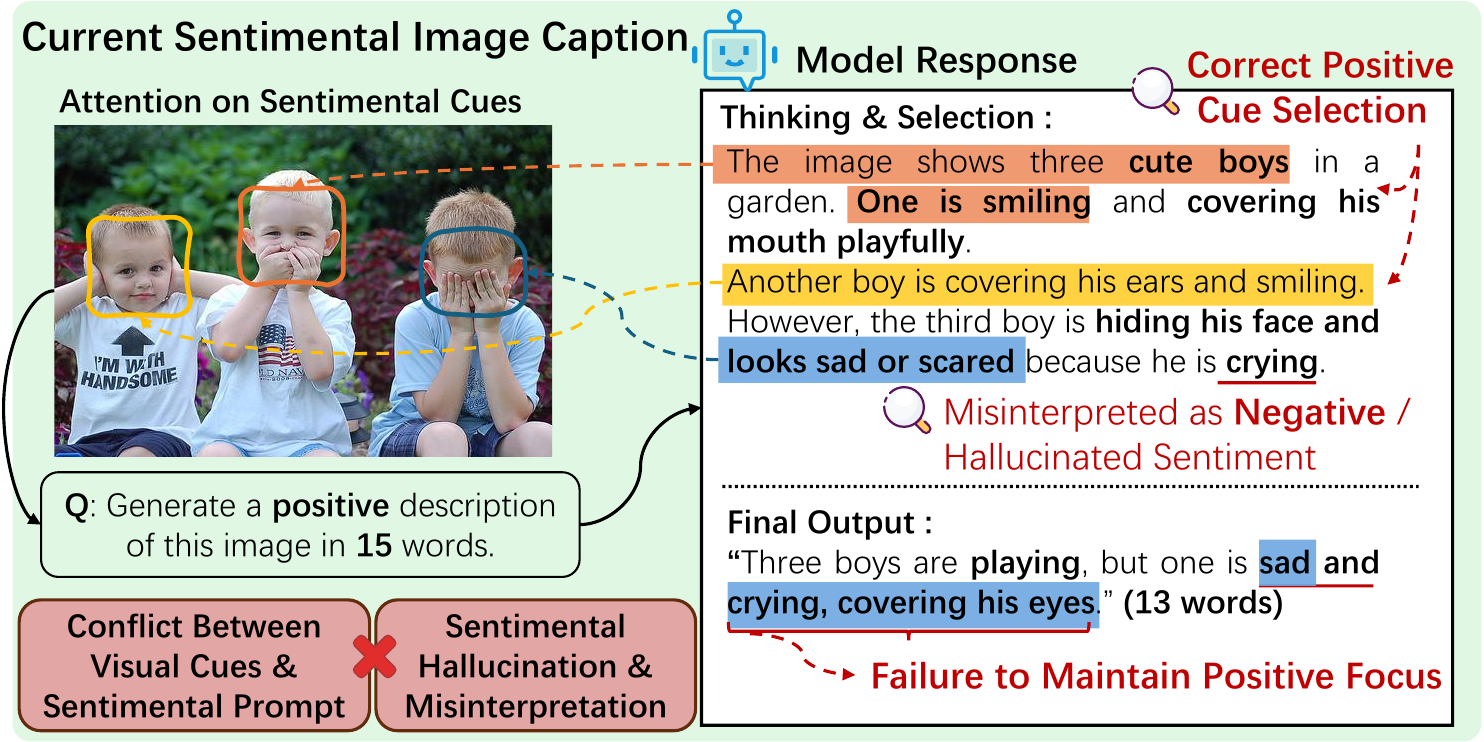}}
\caption{An illustrative failure case of SIC under sentiment and length constraints. Model attends to ambiguous local visual cues and misinterprets or hallucinates negative affect (e.g., \emph{sad}, \emph{crying}), resulting in a caption that violates the intended positive focus. The example highlights the conflict between local visual evidence and sentiment control, and motivates evidence-aware emotion focus selection and verification.}
\label{example}
\vspace{-0.5cm}
\end{figure}

As illustrated in Fig.~\ref{example}, this difficulty is pronounced in scenes containing mixed or ambiguous sentimental cues. The image presents three boys exhibiting contrasting affective signals, where two are smiling while the third covers his face. When the model generates a positive description under a strict length constraint (e.g., 15 words), the reasoning process exhibits a conflict. Although the model identifies positive cues from smiling subjects during intermediate processing, it misinterprets the ambiguous gesture of the third boy as crying and introduces a negative hallucination. Due to the limited word budget, the model is forced to compress the content and erroneously prioritizes this salient negative feature over the requested positive sentiment. Consequently, the final output violates the sentiment prompt by focusing on the misinterpreted subject. This failure exemplifies two fundamental limitations in existing SIC methods: (i) \textbf{Ambiguous sentiment-object grounding}: Many approaches treat sentiment as a global image attribute relying on image-level labels. They fail to explicitly localize verifiable sentiment-related focuses on specific objects (e.g., grounding ``joy'' specifically to the smiling boys), leading to confused sentimental interpretations when multiple objects differ in affect. (ii) \textbf{Lack of evidence-based verification}: Most generation pipelines lack a feedback mechanism grounded in visual evidence. Once a negative sentiment is hallucinated (e.g., misinterpreting a gesture as ``crying''), the system has no means to verify this against the input image or correct the drift to align with the prompt. These issues raise a central challenge in SIC: \emph{how to jointly ensure sentimental accuracy and factual consistency at complex constraints?}

To address these two issues, we propose a Sentiment-Evidence-Aware Multi-Agent System for sentimental image captioning (SEA-Cap). 
By mining diverse sentimental cues in an image, we localize sentiment-relevant focuses to improve the accuracy of sentimental expression. Through evidence-based reviewing and structured feedback, we refine sentiment control from a global image attribute to verifiable object-level evidence, thereby mitigating hallucinations in sentimental caption generation.
Specifically, we first employ a MLLM as the \emph{Generator} to produce candidate captions conditioned on the target sentiment and length constraints. 
Next, an \emph{Sentiment Evidence Miner} performs target-aware segmentation and local affective analysis to extract sentiment-relevant regions, and to construct structured visual sentiment evidence. 
Based on the mined evidence, an evidence-driven \emph{Hallucination Checker} audits candidate captions by verifying object existence, sentiment-focus plausibility, and relation/context consistency, and to write structured feedback into a shared blackboard memory. 
Finally, an \emph{Arbitrator} reads the blackboard and determines whether to accept the current caption or to trigger another revision round, enabling iterative refinement toward sentiment-consistent and fact-faithful descriptions. 
We summarize our main contributions as follows:
\begin{itemize}
    \item We propose a sentiment-evidence-aware multi-agent framework for SIC, which integrates a multimodal generator, evidence miner, evidence-based verifier, and arbitrator through a shared blackboard for iterative refinement. 
    \item We introduce an Sentiment Evidence Miner that extracts target-aware local affective regions via segmentation and local sentiment analysis, producing structured visual sentiment evidence to support verifiable sentiment control.
    \item We design an
    Hallucination Checker that leverages mined sentiment evidence to verify object existence, sentiment-focus rationality, and relation/context consistency, providing structured feedback to improve factual grounding.
    \item Extensive experiments on two widely used benchmark datasets demonstrate that our proposed method achieves state-of-the-art performance on the SIC task.
\end{itemize}

\section{Related Work}
\label{relatedwork}
We briefly review related studies from SIC and MAS for image captioning. 
These lines summarize the progress from early sentiment models to recent MLLMs, and motivate the necessity for reliable multi-objective coordination.

\subsection{Sentimental Image Captioning}
Sentimental Image Captioning (SIC) extends factual captioning by requiring captions to express a target sentiment while remaining visually faithful. Early methods achieved this by injecting emotional signals via specialized embeddings\cite{Senticap,Stylenet,Semstyle} or adaptive attention mechanisms \cite{Guo2019Mscap}.
Driven by the evolution of MLLMs \cite{li2023blip,bai2025qwen2,meta2024llama}, recent approaches have shifted toward large-scale pretraining \cite{Concap} and prompt-based zero-shot refinement \cite{Conzic} to enhance controllability. However, integrating length constraints into SIC remains a significant challenge. Although existing techniques ranging from conditioned decoders to post-hoc rewriting\cite{deng2020length,GuLength,RetkowskiW25Length} can satisfy word budgets, their direct application to SIC compromises affective coherence and visual grounding. Furthermore, coordinating the competing objectives of sentiment, length, and fidelity remains an open problem that single-agent strategies struggle to resolve.

\subsection{Multi-Agent System for Image Captioning}
MASs decompose complex generation into cooperative subtasks, improving diversity, factual consistency, and contextual alignment \cite{bai2025power,lee2025toward,shen2023hugginggpt}.
MosAIC~\cite{bai2025power} uses multiple agents to represent cultural views with a central aggregator. 
MASs have also been explored for grounding and hallucination reduction: CapMAS~\cite{lee2025toward} coordinates planner, generator, and critic agents to enhance factual completeness, and Jiang~et~al.~\cite{JIANG2025114977} combine relational reasoning with retrieval-augmented generation and hallucination detection to strengthen entity-level grounding.
Follow this, IE-MAS integrates internal representation-level steering with external multi-agent collaboration for controllable IC \cite{cai2025ie}.
Nevertheless, effective coordination remains difficult due to conflicts among potentially competing objectives, which may further induce hallucination \cite{yang2019image}.
\begin{figure*}[ht]
    \centering
    
    \includegraphics[width=0.95\textwidth]{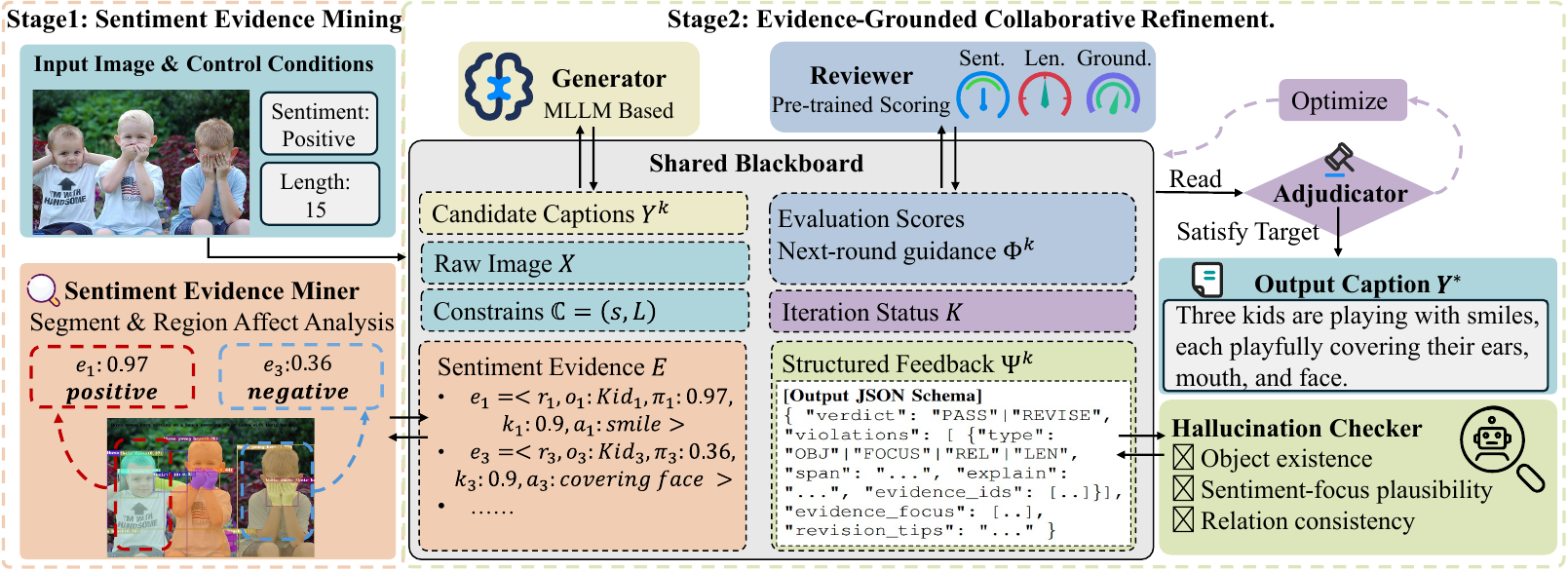}
    \caption{Overview of the SEA-Cap framework. \textbf{Sentiment Evidence Mining} extracts sentimental visual evidence $E$. \textbf{Collaborative Refinement} involves the \textit{Generator}, \textit{Reviewer}, \textit{Hallucination Checker} and \textit{Adjudicator} interacting through a Shared Blackboard. This multi-agent system iteratively optimizes candidate captions $\mathbf{Y^k}$  using structured feedback $\Psi^k$  to ensure the final output $\mathbf{Y^*}$  satisfies target constraints $\mathbb{C}$.}
    
    \label{fig:overall}
\vspace{-0.5cm}
\end{figure*}
\section{Preliminary}
\textbf{Sentimental Image Captioning (SIC)} aims to generate a caption that is \textit{factually grounded} in an input image $X$ while complying with controllable constraints, such as a target sentiment $s$  and a word-budget $L$.
The image $X$ is encoded by a vision encoder $\phi$  into a visual representation $v=\phi(X)\in\mathbb{R}^d$, $s\in\mathcal{S}$ specifies the desired sentiment polarity (e.g., positive or negative) and $L\in\mathbb{Z}^+$ denotes the target length.
Given a generative model $G$, the caption is produced as a token sequence $Y=\{y_1,y_2,\ldots,y_T\}$ and follows the standard autoregressive factorization \cite{chen2020say}:
\begin{equation}
\label{eq:autoreg}
\Pr(Y \mid \mathbb{C})=\prod_{t=1}^{T} \Pr\left(y_t \mid y_{<t},\, s,\, L,\, v\right),
\end{equation}
where we consider the constraint set $\mathbb{C}=\{L,s,v\}$ to capture their joint effects on generation.

\textbf{Multi-Agent System (MAS)} provides an iterative collaboration framework for satisfying various constraints for SIC tasks. 
A minimal MAS consists of a \emph{Generator} generating candidate captions, a \emph{Reviewer} providing diagnostic feedback, and an \emph{Adjudicator} decides acceptance or selects the best candidate under a maximum budget $K_{\max}$.
Specifically,
at iteration $k$,
the Generator synthesizes a candidate caption $\mathbf{Y}^{(k)}$ by integrating the image $X$, constrains $\mathbb{C}$, and the preceding guidance $\Phi^{(k-1)}$. For the initial step, $\Phi^{(0)}$ is defined as the foundational prompt to seed the generation.
Then, 
the Reviewer outputs diagnostic scores for length, sentiment, and factual grounding, together with a brief rationale $\theta^{(k)}$, and summarizes them into the next-round guidance $\Phi^{(k)}$.
Finally,
the Adjudicator determines whether to accept $\mathbf{Y}^{(k)}$ based on the Reviewer's feedback, and selects the optimal $\mathbf{Y}^{*}$ from all candidates if reaching maximal iterations.

\section{Methodology}
To achieve accurate sentimental expression while maintaining factual grounding, we introduce SEA-Cap, a sentiment-evidence-aware multi-agent framework for SIC. 
SEA-Cap is built on the insight that while MLLMs generate fluent captions, reliable control necessitates image-grounded sentiment cues and iterative coordination to reconcile conflicts between sentiment, length, and grounding.
As shown in Fig.~\ref{fig:overall}, SEA-Cap operates in two coupled stages. 
\textbf{Stage~1: Sentiment Evidence Mining.} Given an input image, the Sentiment Evidence Miner performs target-aware segmentation and local affective analysis to extract sentiment-relevant regions, and summarizes them as structured visual evidence. 
\textbf{Stage~2: Evidence-Grounded Collaborative Refinement.} Guided on the control conditions and the mined evidence, the Generator produces an initial draft and posts it to a shared blackboard.
The Hallucination Checker then verifies the draft by checking object existence, sentiment-focus alignment, and contextual consistency, and returns structured feedback to the blackboard. 
Finally, the Adjudicator aggregates the blackboard states to decide whether to accept the caption or to issue refined instructions for another revision round, yielding a verified sentimental caption.

\subsection{Sentiment Evidence Mining.}

The \emph{Sentiment Evidence Miner} $\mathcal{M}$ is designed to decompose the input image into \textbf{verifiable, object-level} sentiment evidence. This structured output enables downstream agents to validate whether a generated caption anchors its affective expression in substantiated visual cues.
Given an image $X$, $\mathcal{M}$ outputs a structured evidence set
$E=\{e_i\}_{i=1}^{N}$ and writes into the shared blackboard $\mathcal{B}$.
Each evidence item $e_i$ corresponds to an object-level region and its local affect cues:
\begin{equation}
e_i = \langle r_i, o_i, \pi_i, \kappa_i, a_i \rangle,
\label{eq:evidence_tuple}
\end{equation}
where $r_i$ denotes the region mask, $o_i$ is the object label, $\pi_i$ is the \emph{sentiment relevance} score, $\kappa_i$ is the \emph{local affect confidence}, and $a_i$ stores optional affect attributes (e.g., ``smiling'', ``covering face'', facial expression, posture cues), which can be extracted by off-the-shelf attribute or expression analyzers.

\textbf{Target-aware segmentation.}
We first generate candidate object-level regions using a generic segmentation module $\mathcal{S}$:
\begin{equation}
R=\{r_i\}_{i=1}^{N} = \mathcal{S}(X).
\label{eq:segmentation}
\end{equation}
To better align regions with the \emph{target} sentiment $s$, we build a sentiment-related query set $\mathcal{Q}(s)$ and score each region by region-query compatibility.
Let $\mathbf{z}^v_i=f_v(X\odot r_i)$ be the visual embedding of region $r_i$ (masked crop), and $\mathbf{z}^t_q=f_t(q)$ be the text embedding of a query $q \in \mathcal{Q}(s)$.
We define a sentiment-aware saliency score:
\begin{equation}
\gamma_i \;=\; \max_{q \in \mathcal{Q}(s)} \mathrm{sim}(\mathbf{z}^v_i,\mathbf{z}^t_q).
\label{eq:saliency}
\end{equation}
Then we keep the top-$K$ regions, which encourages the miner to prioritize objects contributing to the target emotion.

\textbf{Local affective analysis.}
For each candidate region $r_i$, an affect analyzer $A$ estimates the local affective intensity by:
\begin{equation}
p_i = A(X\odot r_i) \in \Delta^{|\mathcal{S}|},
\end{equation}
\begin{equation}
\hat{s}_i = \arg\max_{s' \in \mathcal{S}} p_i[s'] ,
\end{equation}
where $p_i$ is the predicted sentiment distribution over $\mathcal{S}$, and $\hat{s}_i$ is the most likely local sentiment label.

\textbf{Blackboard evidence writing.}
Finally, $\mathcal{M}$ aggregates these analysis results and registers $E$ onto the shared blackboard $\mathcal{B}$:
\begin{equation}
\mathcal{B}\leftarrow \mathcal{B}\cup\{E\}.
\end{equation}
This registration shifts sentiment control from coarse image-level labeling to fine-grained, object-level evidence grounding. By providing a verifiable basis for the subsequent \emph{Hallucination Checker}, this evidence-centric memory supports iterative refinement, allowing the system to mitigate sentiment drift and hallucinated affect even under strict length budgets.

\subsection{Evidence-Grounded Collaborative Refinement.}

The \emph{Hallucination Checker} $\mathcal{H}$ performs evidence-grounded and verifiable auditing for each candidate caption.
Different from a purely text-based self-critique, $\mathcal{H}$ conditions its judgment on the structured sentiment evidence mined by $\mathcal{M}$, thereby enabling object-level verification of sentiment focus and factual consistency.

At iteration $k$, $\mathcal{H}$ reads the shared blackboard $\mathcal{B}^{(k)}$, including the image $X$, control specification $\mathbb{C}=\{s,L\}$, the candidate caption $\mathbf{Y}^{(k)}$, and the evidence set $E^{(k)}$.
$\mathcal{H}$ audits $\mathbf{Y}^{(k)}$ from three complementary aspects:

\textbf{Object existence.}
Let $O_Y$ be the set of object mentions extracted from the caption, 
and $O_E=\{o_i\}$ be the evidence object set.
We define a mention-to-evidence matching function $m(o)\in\{1,\dots,N\}\cup\{\varnothing\}$ that maps a caption mention $o$ to the most semantically compatible evidence object $o_i$. 
Then object existence violations are detected by:
\begin{equation}
V_{\text{obj}} = \{\, o \in O_Y \mid m(o)=\varnothing \,\},
\end{equation}
which indicates hallucinated entities not supported by the mined evidence.

\textbf{Sentiment-focus plausibility.}
Let $T_Y$ denote sentiment-bearing spans in the caption, e.g. adjectives or affect verbs.
Since each evidence item $e_i=\langle r_i,o_i,\pi_i,\kappa_i,a_i\rangle$ provides the sentiment relevance $\pi_i$, we define focus-supported objects as:
\begin{equation}
F(E^{(k)}) = \{\, i \mid \pi_i \ge \tau_{\pi} \,\},
\end{equation}
where $\tau_{\pi}$ is a threshold.
For each sentiment span $t \in T_Y$, $\mathcal{H}$ checks whether $t$ is anchored to a matched object $m(o)$ with $m(o)\in F(E^{(k)})$; otherwise it is deemed \emph{unsubstantiated sentiment focus}.
This prevents the caption from grounding the requested sentiment on ambiguous or low-relevance regions.

\textbf{Relation consistency.}
For a relational statements parsed from the caption $c=(o_a, p, o_b)$, we use evidence matching to locate its supporting items.
We flag a relation as unsupported if any endpoint is missing, or the predicate $p$ contradicts the evidence attributes $\mathbf{a}_{i}$ or the visual context implied by regions.

\textbf{Structured feedback format.}
We implement $\mathcal{H}$ as an MLLM-based agent that is constrained to output machine-readable feedback.
Given $\langle X, \mathbb{C}, \mathbf{Y}^{(k)}, E^{(k)} \rangle$, we prompt the model with evidence items and require a strict JSON output $\Psi^{(k)}$.
This structured design makes the feedback directly writable into the blackboard and readily consumable by the Arbitrator for iterative refinement.

\begin{table*}[t]
\definecolor{hlcolor}{rgb}{0.92, 0.95, 0.99}
\centering
\caption{Performance comparison on SentiCap and FlickrStyle datasets across different sentiments. The best results in each category are highlighted in \textbf{bold}. $\downarrow$ indicates lower is better.}
\label{tab:mainresults}
\resizebox{0.76\linewidth}{!}{
\begin{tabular}{lllccccc}
\toprule
Dataset & Sentiment & Model & MAE $\downarrow$ & Cls. $\uparrow$ & ClipScore $\uparrow$ & $\text{CHAIR}_i \downarrow$ & $\text{CHAIR}_s \downarrow$ \\ 
\midrule
\multirow{12}{*}{SentiCap} & \multirow{6}{*}{Positive} & PositionID & 10.203 & 0.562 & 0.681 & 0.231 & 0.437 \\
& & ConZIC     & 1.632  & 0.477 & 0.799 & 0.208 & 0.212 \\
& & Qwen2.5-VL & 3.500  & 0.933 & 0.773 & 0.172 & 0.181 \\
& & LLaMA3.2-v & 1.942  & 0.827 & 0.802 & 0.169 & 0.183 \\
& & IE-MAS     & 0.785  & 0.958 & 0.812 & 0.152 & 0.158 \\
& &  \textbf{SEA-Cap} & \textbf{0.732} & \textbf{0.962} & \textbf{0.838} & \textbf{0.128} & \textbf{0.131} \\ 
\cmidrule{2-8}
& \multirow{6}{*}{Negative} 
& PositionID & 12.274 & 0.498 & 0.648 & 0.253 & 0.457 \\
& & ConZIC     & 1.636  & 0.722 & 0.801 & 0.211 & 0.242 \\
& & Qwen2.5-VL & 3.161  & 0.425 & 0.798 & 0.157 & 0.231 \\
& & LLaMA3.2-v & 1.932  & 0.917 & 0.792 & 0.163 & 0.256 \\
& & IE-MAS     & 0.773  & 0.928 & 0.804 & 0.161 & 0.198 \\
& &  \textbf{SEA-Cap} & \textbf{0.746} & \textbf{0.953} & \textbf{0.822} & \textbf{0.136} & \textbf{0.166} \\ 
\midrule
\multirow{8}{*}{FlickrStyle} & \multirow{4}{*}{Romantic} 
& Qwen2.5-VL & 3.833  & 0.300 & 0.386 & 0.531 & 0.683 \\
& & LLaMA3.2-v & 3.243  & 0.470 & 0.550 & 0.496 & 0.526 \\
& & IE-MAS     & 1.953  & 0.733 & 0.393 & 0.438 & 0.501 \\
& &  \textbf{SEA-Cap} & \textbf{1.625} & \textbf{0.751} & \textbf{0.804} & \textbf{0.288} & \textbf{0.389} \\ \cmidrule{2-8}
& \multirow{4}{*}{Humorous} 
& Qwen2.5-VL & 3.945  & 0.140 & 0.772 & 0.323 & 0.391 \\
& & LLaMA3.2-v & 3.375  & 0.279 & 0.782 & 0.387 & 0.406 \\
& & IE-MAS     & 1.763  & 0.402 & 0.784 & 0.322 & 0.392 \\
& &  \textbf{SEA-Cap} & \textbf{1.533} & \textbf{0.571} & \textbf{0.813} & \textbf{0.298} & \textbf{0.327} \\ 
\bottomrule
\end{tabular}
}
\vspace{-0.3cm}
\end{table*}

\subsection{Shared Blackboard \& Community Patrol}
\textbf{Shared blackboard.}
SEA-Cap adopts a \emph{shared blackboard} as a centralized, structured workspace that stores the immutable inputs and accumulates intermediate artifacts across agents.
At iteration $k$, we represent the blackboard state as:
\begin{equation}
\mathcal{B}^{(k)} 
= \Big\{
X, \mathbb{C}, E^{(k)},
\mathbf{Y}^{(k)},
\Psi^{(k)},
\Phi^{(k)}
\Big\},
\label{eq:blackboard_state_protocol}
\end{equation}
where inputs $X$, $\mathbb{C}$ remain fixed, while $E^{(k)}, \mathbf{Y}^{(k)}, \Psi^{(k)}, \Phi^{(k)}$ are iteratively updated.

\textbf{Communication protocol.}
Instead of ad-hoc natural language exchanges between agents, SEA-Cap enforces a unified \emph{protocol} over $\mathcal{B}$.
Each agent reads only the required fields and writes back \emph{typed, machine-readable} messages.
This blackboard-centric protocol offers advantages that:
(i) each refinement step is backed by explicit evidence and typed feedback records.
(ii) agents communicate through a consistent schema, reducing error propagation caused by free-form dialogues.
(iii) each component can be replaced by another off-the-shelf model without changing the overall loop, facilitating fair comparisons across backbones.

\section{Experiment}
\subsection{Experimental~Settings}
\textbf{Datasets.}
We evaluate the performance of SEA-Cap on
two benchmark datasets: SentiCap and FlickrStyle.
\textsc{SentiCap}~\cite{Senticap} contains 2000 images, each annotated with three human-written captions expressing either positive or negative~sentiment. \textsc{FlickrStyle}~\cite{Stylenet} has 10,000 pairs of images and stylized captions including humorous and romantic sentiment.

\textbf{Metrics.}
Following standard evaluation protocols \cite{cai2025ie}, we adopt \textit{Mean Absolute Error (MAE)}, \textit{Classification Accuracy (Cls.)} and \textit{CLIPScore} to evalute length controlability, sentiment expression and image-semantic alignment, respectively.
Regarding hallucination, we utilize the \textit{CHAIR} metric \cite{rohrbach-etal-2018-object} to evaluate object consistency. Specifically, we report \textit{CHAIR$_i$} and \textit{CHAIR$_s$}, which denote the proportion of hallucinated object instances and the proportion of sentences containing at least one hallucinated object, respectively.
By default, the length constraint $L=15$ and the maximum iteration is 5. All experiments are executed on NVIDIA A800~GPU.

\subsection{Analysis of Main Results}
\textbf{Performance Comparison on SentiCap.} As illustrated in Table \ref{tab:mainresults}, the proposed SEA-Cap framework achieves a superior balance between sentiment adherence and descriptive accuracy compared to state-of-the-art baselines. On the SentiCap dataset, which features canonical positive and negative sentiments, our model consistently outperforms both optimization-based methods (e.g., ConZIC) and large-scale multimodal models (e.g., Qwen2.5-VL) across all metrics.
A key observation is that while many baselines struggle to satisfy the target length constraint, SEA-Cap reduces length deviation to below 0.8 in both settings. This precision suggests that the iterative feedback loop between the Judge and the Generator effectively enforces the word budget. Furthermore, our model significantly reduces object hallucination, achieving the lowest $CHAIR_i$ and $CHAIR_s$ scores (e.g., 0.128/0.131 for Positive). This demonstrates that the Hallucination Checker, guided by structured evidence from the Sentiment Evidence Miner, prevents the model from fabricating entities to satisfy sentimental requirements.

\textbf{Generalization to Complex Sentiments on FlickrStyle.} The advantage of the multi-agent system becomes even more evident on the FlickrStyle dataset (Romantic and Humorous). Unlike basic polar sentiments, these sentiments rely on subtle, context-dependent cues. 
As shown in the bottom half of Table \ref{tab:mainresults}, the baseline methods suffer from poor sentiment classification accuracy (Cls.). In contrast, SEA-Cap maintains high sentiment fidelity (0.751 for Romantic) and superior image-text alignment (0.804 CLIPScore). This improvement highlights that our framework’s ``Shared Blackboard" mechanism effectively integrates local sentimental evidence into the global captioning process, enabling the model to capture complex sentiment cues without compromising visual grounding.

\textbf{Baseline Methods.}
Our SEA-Cap is compared against two backbone MLLMs: \textit{Qwen2.5-VL (7B)} \cite{bai2025qwen2} and \textit{LLaMA 3.2 Vision (11B)} \cite{meta2024llama} (without our multi-agent control). 
Additionally, representative control-oriented methods are included for comparison: \textit{ConZIC} \cite{Conzic}, a~Gibbs-sampling-based controllable captioner, \textit{PositionID} \cite{Positionid}, a~positional-prompting method for explicit length~regulation and \textit{IE-MAS} \cite{cai2025ie}, an MAS-based method for multiple constraints captioner.

\textbf{Implementation Details.}
We build our framwork based on the \texttt{LLaMA3.2-vision (11B)} \cite{meta2024llama}. We adopt the \texttt{SAM} \cite{Kirillov_2023_ICCV} as the base segmentation backbone for the Sentiment Evidence Miner, and use \texttt{CLIP} as a lightweight, training-free proxy for local affect cues.

\begin{table}[t]
\centering
\caption{Ablation study of our proposed SEA-Cap model on SentiCap dataset. ``w/o $\mathcal{M}$'' and ``w/o $\mathcal{H}$'' denote the model without the \textbf{Sentiment Evidence Miner} and the \textbf{Hallucination Checker}, respectively. Bold values indicate the best performance.}
\label{tab:ablation}
\setlength{\tabcolsep}{2pt} 
\begin{tabular}{llccccc}
\toprule
Model & Sent. & MAE $\downarrow$ & Cls. $\uparrow$ & ClipS. $\uparrow$ & $\text{CHAIR}_i \downarrow$ & $\text{CHAIR}_s \downarrow$ \\ \midrule
\textbf{SEA-Cap} & \multirow{3}{*}{Pos} & \textbf{0.732} & \textbf{0.962} & \textbf{0.838} & \textbf{0.128} & \textbf{0.131} \\
w/o $\mathcal{M}$            &                      & 0.801          & 0.915          & 0.812          & 0.138          & 0.139          \\
w/o $\mathcal{H}$            &                      & 1.833          & 0.821          & 0.798          & 0.197          & 0.221          \\ \midrule
\textbf{SEA-Cap} & \multirow{3}{*}{Neg} & \textbf{0.746} & \textbf{0.953} & \textbf{0.822} & \textbf{0.136} & \textbf{0.166} \\
w/o $\mathcal{M}$            &                      & 0.931          & 0.892          & 0.822          & 0.139          & 0.142          \\
w/o $\mathcal{H}$            &                      & 1.927          & 0.857          & 0.785          & 0.203          & 0.217          \\ \bottomrule
\end{tabular}
\vspace{-0.5cm}
\end{table}

\begin{figure*}[t]
\centering
\includegraphics[width=0.95\textwidth]{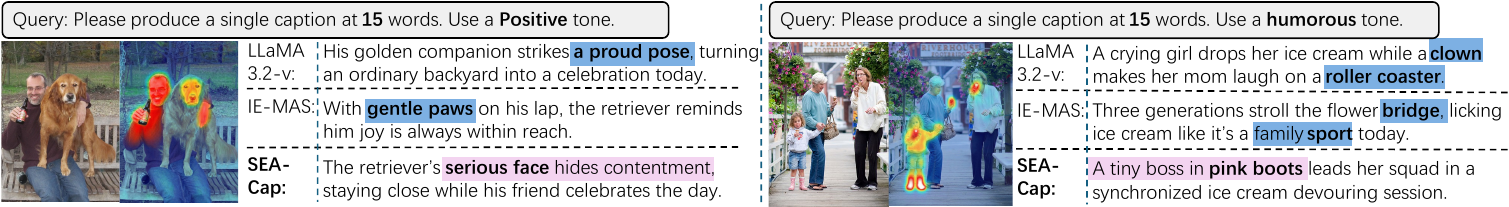}
\caption{Qualitative comparison of sentiment captioning results under length constraints. The examples demonstrate \textbf{Positive} (left) and \textbf{Humorous} (right) sentiment targets. The visualized heatmaps illustrate that our method explicitly focuses on local regions to mine sentiment evidence. Text in \textcolor{blue}{blue} indicates hallucinations or inconsistencies with visual evidence, while text in \textcolor{magenta}{pink} highlights accurate grounding that enhances the target sentiment. 
}
\label{case}
\vspace{-0.5cm}
\end{figure*}

\subsection{Ablation Study}
To evaluate the contribution of each core component in our SEA-Cap, we conduct an ablation study on the SentiCap dataset. As summarized in Table \ref{tab:ablation}, we examine the performance of SEA-Cap by systematically removing the Sentiment Evidence Miner ($\mathcal{M}$) and the Hallucination Checker ($\mathcal{H}$). 

\textbf{Impact of Sentiment Evidence Miner ($\mathcal{M}$).} Removing $\mathcal{M}$ leads to a noticeable decline in sentiment classification accuracy ($Cls.$) across both positive and negative settings. 
This suggests that local, object-level sentiment evidence is crucial for grounding affective descriptions. Without this agent, the system relies more on global, image-level attributes, which reduces its sentimental precision.

\textbf{Impact of Hallucination Checker ($\mathcal{H}$).} The removal of $\mathcal{H}$ results in the most significant performance degradation across all metrics, particularly in CHAIR scores. Specifically, the hallucination rates ($CHAIR_i$) nearly double from 0.136 to 0.203 in negative settings. 
These findings underscore that without structured feedback and evidence-grounded checking, the generator struggles to adhere to strict word budgets and factual consistency. $\mathcal{H}$ acts as a critical corrective force that coordinates competing objectives and prevents the accumulation of semantic hallucinations. The results of this ablation study demonstrate that the synergy between evidence mining and structured checking is essential for achieving high-fidelity.

\subsection{Qualitative Comparison}
Fig. ~\ref{case} presents qualitative comparisons of sentiment captioning quality under length and sentiment constraints. 
In the Positive case, SEA-Cap captures the subtle ``serious face" of the dog to convey a unique ``contentment."  Similarly, in more complex Humorous scenario, baselines struggle to generate amusement without sacrificing truthfulness. For example, LLaMA 3.2-v fails catastrophically by hallucinating a ``clown" and ``roller coaster" to force a funny narrative. Conversely, SEA-Cap grounds its output in the "pink boots" and the child's expression as key evidence. The "tiny boss" metaphor, achieving humorous through factual re-contextualization rather than fabrication. These results confirm that our evidence-aware mechanism effectively prevents the sentiment drift and object hallucinations prevalent in existing methods.
\section{Conclusion}
In this paper, we introduce SEA-Cap, a Sentiment-Evidence-Aware Multi-Agent System for SIC. Specifically, we first propose the Sentiment Evidence Miner, which performs local affective analysis to extract and structure sentiment-relevant visual evidence from the input image, resolving the ambiguity of global sentiment attributes. 
Furthermore, we present an evidence-driven multi-agent collaboration mechanism, incorporating a Generator, Reviewer, Hallucination Checker, and Adjustor. This framework utilizes a shared blackboard to iteratively audit and refine candidate captions against mined evidence, thereby ensuring both sentimental precision and factual consistency. 
Extensive experiments and ablation studies on two widely used benchmarks validate the effectiveness of our proposed method and demonstrate its superiority in mitigating hallucinations for the SIC task.

{
\let\oldbibliography\thebibliography
\renewcommand{\thebibliography}[1]{
\oldbibliography{#1}
\scriptsize
}
\bibliographystyle{IEEEbib}
\bibliography{main}
}

\end{document}